\documentclass{article}

\usepackage[final]{nips_2018}

\usepackage{color}

\usepackage[utf8]{inputenc} 
\usepackage[T1]{fontenc}    
\usepackage{hyperref}       
\usepackage{url}            
\usepackage{booktabs}       
\usepackage{amsfonts}       
\usepackage{nicefrac}       
\usepackage{microtype}      

\title{Environments for Lifelong Reinforcement Learning}

\author{
  Khimya Khetarpal\thanks{Equal contribution} \\
  Mila, McGill University, Montreal, Canada\\
  \texttt{khimya.khetarpal@mail.mcgill.ca} \\
  \And
   Shagun Sodhani\footnotemark[1] \\
   Mila, Universit\'{e} de Montr\'{e}al, Canada \\
   \texttt{sshagunsodhani@gmail.com} \\
  \AND
   Sarath Chandar \\
   Mila, Universit\'{e} de Montr\'{e}al, Canada \\
  \texttt{sarathcse2008@gmail.com} \\
  \And
  Doina Precup \\
  Mila, McGill University, Montreal, Canada \\
  \texttt{dprecup@cs.mcgill.ca} \\
}

\begin{document}

\maketitle

\begin{abstract}
  To achieve general artificial intelligence, reinforcement learning (RL) agents should learn not only to optimize returns for one specific task but also to constantly build more complex skills and scaffold their knowledge about the world, without forgetting what has already been learned. In this paper, we discuss the desired characteristics of environments that can support the training and evaluation of lifelong reinforcement learning agents, review existing environments from this perspective, and propose recommendations for devising suitable environments in the future.
\end{abstract}

\section{Introduction}

Humans acquire skills and build on them to solve increasingly complex tasks. For instance, consider a child learning to play basketball. This involves learning to hold the ball properly, to throw and catch, then learning to pass and dribble. These skills are then combined to learn more complex skills. For example, a lay-up is a composition of dribbling while running and a throw, while ensuring that the sequence of steps is not too long and that the throw lands the ball in the hoop. The entire task of playing a game relies on various subtasks and requires developing skills of increasing complexity. More importantly, skills need to work in different games and against different opponents.

In contrast, reinforcement learning (RL) agents, while able to achieve human-level performance in complex games like Go, usually focus on becoming really proficient at one task, and train from scratch in each new problem they face.

Lifelong learning agents can learn from a stream of experience spanning many tasks (possibly of different nature) over its lifetime \citep{silver2013lifelong,ring1997child,thrun1996explanation}.  The early work of Ring \citep{ring1997child} describes a continual learning agent as an autonomous agent trained on a sequence of tasks with no final task. This process bears different names in the  literature - incremental and continual learning \citep{solomonoff1989system}, never-ending learning \citep{carlson2010toward}, etc.

In this work, we focus on lifelong learning performed in the context of reinforcement learning.  In addition to optimizing returns,  a lifelong RL agent should be able to:
\begin{itemize}
\setlength\itemsep{-0.1em}
    \item Learn behaviors, skills, and predictions about the environment while solving given tasks.
    \item Learn incrementally throughout its lifetime
    \item Combine previously learned skills and build on top of them to solve increasingly complex tasks
    \item Plan for short-term and long-term goals
\end{itemize}

We argue that virtual-embodiment is perhaps the most natural setup for training and evaluating lifelong reinforcement learning agents. We review existing environments to understand if they are appropriate for training and evaluating a lifelong learning agent. Based on this comparative analysis, we propose recommendations for RL environments suitable for lifelong learning.

\section{Background and Motivation} 
Recent breakthroughs in RL research \citep{silver2016mastering, mnih2015human} have been powered in part by advances in deep learning and the availability of diverse simulation environments to train RL agents. The Arcade Learning Environment (ALE) \citep{bellemare2013arcade}, originally proposed in $2013$, is a suite of Atari $2600$ games which provides dozens of problems in which to train and evaluate RL agents. More recently, OpenAI's Gym \citep{brockman2016openai} was developed and offers a broader variety of environments ranging from toy text and grid world problems to continuous control tasks and Atari-based games. 

One of the shortcomings of the originally proposed ALE platform is the deterministic nature of the environments, which can result in memorization of state-action sequences as opposed to generalization. A more recent version of ALE \citep{machado2018revisiting} supports multiple game modes and provides a form of stochasticity. 

Moving towards a more realistic setting, frameworks such as DeepMind Lab \citep{beattie2016deepmind} and VizDoom \citep{kempka2016vizdoom} offer $3$D first-person-view environments. Both VizDoom and DeepMind Lab support stylized labyrinths catered to navigation tasks. However, they lack naturalistic appeal in terms of their layout, appearance and objects. Another platform that the RL community has explored in recent years is Minecraft \citep{duncan2011minecraft}, which offers a highly complex environment with characteristics that could be potentially leveraged for lifelong learning \citep{tessler2017deep}.

Simulators such as Gazebo \citep{koenig2004design} have been intensively used in robotics research and are tailored towards learning agents which train in simulation but are then evaluated on physical robots. Aligned with recent efforts toward reproducible research, \citep{mahmood2018benchmarking} introduced benchmark tasks for physical robots which allow experiments to be reproduced in different locations and under diverse conditions. While these tasks are quite challenging and push RL agents' limits, for lifelong learning they may be too difficult at the moment, given that robotics platforms have limitations in terms of their ability to support many different tasks. However, they do have a feature which we find very useful for continual learning: embodiment. We will discuss next what we mean by embodiment and why it is useful for lifelong learning.

\section{Learning in Embodied Agents}

 Theories of \textit{embodied cognition} \citep{neuro_virtual_embodiment} suggest that cognition is grounded in perception and action. Embodied learning allows agents to actively interact with the environment and utilize a rich, multi-modal sensori-motor stream of data. However, training physically embodied agents can be slow, expensive and sometimes impractical. Therefore, virtual embodiment can be seen as an alternative approach. More importantly, virtual embodiment is closer to how humans learn through interaction with their environment via multiple sensors and effectors of different types.
Interaction modalities include  (3D) vision, audio signals, natural language, experiencing and exerting physical forces etc. Environments that support a multitude of these modalities are generally more difficult to solve, but they provide diversity and richness that can be very useful in order to build good generalizations.

Virtual embodiment has several other advantages such as i) \textbf{curriculum learning} - virtual environments are easy to modify in terms of complexity thereby making it easier to train agents in progressive fashion, ii) \textbf{short-term and long-term goals} -  these are equivalent respectively with skills and  composition of skills, iii) \textbf{mimic agents in real-world scenarios} - embodied learning tries to encapsulate the real world dynamics in a simulated environment as faithfully as possible, which creates more realistic domains and iv) \textbf{cause-and-effect learning} - rich, multi-modal data streams can help agents to understand the causality relationships of various events  and opportunities associated with each object, through actions that are afforded by these objects.

So far, virtual embodiment environments have been used for tasks like navigation \citep{House3D, embodied_navigation_agent}, visual question answering \citep{embodiedqa}, teaching to execute instructions or programs \citep{virtual_home}. We will now summarize the characteristics of these environments, which we believe would provide good testbeds for training and evaluating lifelong learning algorithms.

\section{Virtual Embodiment Environments: A Short Review}

\begin{subsection}{House 3D}
House3D \citep{House3D} is a realistic and extensible environment built on top of the SUNCG dataset \citep{suncgsong2016} (a large dataset of over 45,000 human designed 3D house layouts). The environment supports rendering photo-realistic 3D visuals with support for diverse 3D objects and layouts. Each scene is annotated with 3D coordinates and other meta-data like room and object type. The paper introduced the room navigation task where a set of episodic environments $E = {E_1, .., E_n}$ and a set of semantic concepts $I = {I_1, .., I_m}$ are pre-defined. During each episode, the agent interacts with one of the environments $e \in E$ and is given a concept $i \in I$. The agent starts at a random position in $e$ and at each time-step, receives a visual signal $x_t$ corresponding to the first person view. The agent needs to navigate to reach a target destination. 

While the environment can be customized for defining new tasks and can be used to load other 3D scene datasets (like \citep{matterport}, \citep{Stanford3D} etc), the environment itself does not allow defining varying difficulty tasks. In particular, once a layout is selected, we can not make the layout ``harder'' or ``easier'' for the agent. For instance, one can not ``add'' obstacles that the agent needs to overcome before the task is considered complete.
\end{subsection} 

\begin{subsection}{HoME: Household Multimodal Environment}
HoME\citep{HoME} is similar to House3D as it is also built on top of the SUNCG dataset. Along with 3D visual rendering and semantic image segmentation, HoME also provides natural language descriptions of objects and audio rendering. It further supports rigid body dynamics (through a physics engine) and external forces like gravity. This environment supports many more modalities compared to the other virtual embodiment environments and also supports ``adding'' or ``removing'' objects. This makes it a suitable candidate for training and evaluating lifelong learning agents in RL.
\end{subsection}

\begin{subsection}{MINOS: Multimodal Indoor Simulator for Navigation in Complex Environments}
MINOS\citep{Minos} is a simulation framework specifically designed for multi-sensory navigation models. It can use layouts from both SUNCG and Matterplot3D. While it is not as rich as some other environments (like HoME or House3D), it allows for easy customization. In particular, MINOS supports material variation (for texture and colors), object clutter variation (where a set of specified categories of objects can be removed), navigation goal specification (where goals can be at arbitrary points in space or can be arbitrary instances of a category), task specification (where the task can be specified through an arbitrary Python function which returns reward signals and episode success or failure at each state).

These characteristics make MINOS a good candidate for lifelong learning. It can be easily used to design tasks with different levels of complexity or to design tasks that require the use of just one skill or composition of skills. These benefits also set it apart from other environments like AI2-THOR \citep{AI2-THOR} which supports only 32 single-room environments or CHALET \citep{CHALET} in which only a small set of discrete actions are supported.
\end{subsection}

\begin{subsection}{VirtualHome: Simulating Household Activities via Programs}
VirtualHome\citep{virtualhome} crowdsourced a dataset of ``programs'' for performing different activities in a house. Most common and atomic (inter)actions were identified and implemented in the Unity3D game engine. The programs and the simulated environment can be used to train an agent to perform household tasks based on language instructions. If we think of each atomic action as a skill, then each program, which is a sequence of atomic actions, can be seen as a composition of skills. This is a major advantage of using VirtualHome - the ready availability of the program dataset (which can be seen as a composition of skills). A disadvantage of this environment, however, is that it does not allow for creating variations of a scene, which is important for designing tasks of varying complexity.
\end{subsection}

Before wrapping up this section, we also consider some of the prominent RGB-D datasets which are useful building blocks for developing environments for virtual embodied agents. The environments discussed earlier use one or more of these datasets.

\begin{subsection}{Matterport3D: Learning from RGB-D Data in Indoor Environments}
Matterport3D\citep{matterport} introduced a large RGB-D dataset of indoors scenes (10,800 panoramic views from 194,400 RGB-D images of 90 building-scale scenes). It includes annotations for surface reconstruction, camera poses and 2D and 3D semantics segmentation. Even though an embodied learning agent is not introduced as part of the task setup, many follow-up works like \citep{matterport-emb} and \citep{Minos} used this work as the starting point for defining a plethora of tasks related to computer vision.
\end{subsection}

\begin{subsection}{SUNCG: A Large 3D Model Repository for Indoor Scenes}
SUNCG \citep{suncgsong2016} is a large scale dataset of richly annotated scenes. 
The dataset contains over 45,000 manually created room and object layouts along with semantic annotations. The dataset was created to learn semantic scene completion, where given a single-view depth map observation, a complete 3D representation, along with semantic labels is generated. Follow-up works like \citep{embodiedqa} and \citep{House3D} used it for training embodied agents. 
\end{subsection}

In the following section, we discuss what is missing in these frameworks in order to perform lifelong learning.

\section{Recommendations for a lifelong learning testbed}
To conclude, we discuss features that, in our view, would be important to support by any framework for  training and evaluation of lifelong learning agents.

\textbf{$1$:}
The proposed testbed should support a multitude of tasks of different difficulty. The lowest level tasks should require only one skill to solve, while tasks more complex tasks should require a composition of skills learned in the previous levels. Tasks could for example be structured in a hierarchy, such that task complexity would increase as the learning agent moves up. Solving a task in the $i+1^{th}$ level should require the agent to solve the task at level $i$ and also learn to compose previously acquired skills. The environment should be able to present the agent with tasks that test its ability to make compositions.

\textbf{$2$:}
A lifelong learning framework would ideally provide easy addition or removal of objects from a scene or a variety of scenes. Such a framework would facilitate the incremental expansion of the data that the agent sees over time. This could be achieved by providing the flexibility to scale up the size of the environment, the quantity of objects with which the agent can interact, etc. 
    
\textbf{$3$:}
Tasks should be generated in the environment in a way that requires the agent to do both short-term and long-term planning. Dealing with many goals that span different time scales would test an agent's capacity to learn different types of knowledge and to generalize across time scales.

\textbf{$4$:}
As the learning agent moves to more complex tasks, the environment should continue to challenge it with previously seen tasks, in order to assess whether the agent can resist catastrophic forgetting.

As discussed above, some of the already existing environments align well with these desiderata. However, more work can be done to expand the set of environments used in lifelong learning. More importantly, assessing the performance of lifelong learning agents and defining the objective that they should optimize is still a problem that has not been tackled much, and is critical for progress in our field. We hope that this paper is at least useful in outlining some of the intuitive desiderata for lifelong learning, and that as a community we can all work to make these more formal and precise.

\small
\bibliography{nips_2018}

\begin{thebibliography}{30}
\providecommand{\natexlab}[1]{#1}
\providecommand{\url}[1]{\texttt{#1}}
\expandafter\ifx\csname urlstyle\endcsname\relax
  \providecommand{\doi}[1]{doi: #1}\else
  \providecommand{\doi}{doi: \begingroup \urlstyle{rm}\Url}\fi

\bibitem[{Anderson} et~al.(2017){Anderson}, {Wu}, {Teney}, {Bruce}, {Johnson},
  {S{\"u}nderhauf}, {Reid}, {Gould}, and {van den Hengel}]{matterport-emb}
P.~{Anderson}, Q.~{Wu}, D.~{Teney}, J.~{Bruce}, M.~{Johnson},
  N.~{S{\"u}nderhauf}, I.~{Reid}, S.~{Gould}, and A.~{van den Hengel}.
\newblock {Vision-and-Language Navigation: Interpreting visually-grounded
  navigation instructions in real environments}.
\newblock \emph{ArXiv e-prints}, November 2017.

\bibitem[{Anderson} et~al.(2018){Anderson}, {Chang}, {Singh Chaplot},
  {Dosovitskiy}, {Gupta}, {Koltun}, {Kosecka}, {Malik}, {Mottaghi}, {Savva},
  and {Zamir}]{embodied_navigation_agent}
P.~{Anderson}, A.~{Chang}, D.~{Singh Chaplot}, A.~{Dosovitskiy}, S.~{Gupta},
  V.~{Koltun}, J.~{Kosecka}, J.~{Malik}, R.~{Mottaghi}, M.~{Savva}, and A.~R.
  {Zamir}.
\newblock {On Evaluation of Embodied Navigation Agents}.
\newblock \emph{ArXiv e-prints}, July 2018.

\bibitem[{Armeni} et~al.(2017){Armeni}, {Sax}, {Zamir}, and
  {Savarese}]{Stanford3D}
I.~{Armeni}, A.~{Sax}, A.~R. {Zamir}, and S.~{Savarese}.
\newblock {Joint 2D-3D-Semantic Data for Indoor Scene Understanding}.
\newblock \emph{ArXiv e-prints}, February 2017.

\bibitem[Beattie et~al.(2016)Beattie, Leibo, Teplyashin, Ward, Wainwright,
  K{\"u}ttler, Lefrancq, Green, Vald{\'e}s, Sadik, et~al.]{beattie2016deepmind}
Charles Beattie, Joel~Z Leibo, Denis Teplyashin, Tom Ward, Marcus Wainwright,
  Heinrich K{\"u}ttler, Andrew Lefrancq, Simon Green, V{\'\i}ctor Vald{\'e}s,
  Amir Sadik, et~al.
\newblock Deepmind lab.
\newblock \emph{arXiv preprint arXiv:1612.03801}, 2016.

\bibitem[Bellemare et~al.(2013)Bellemare, Naddaf, Veness, and
  Bowling]{bellemare2013arcade}
Marc~G Bellemare, Yavar Naddaf, Joel Veness, and Michael Bowling.
\newblock The arcade learning environment: An evaluation platform for general
  agents.
\newblock \emph{Journal of Artificial Intelligence Research}, 47:\penalty0
  253--279, 2013.

\bibitem[Brockman et~al.(2016)Brockman, Cheung, Pettersson, Schneider,
  Schulman, Tang, and Zaremba]{brockman2016openai}
Greg Brockman, Vicki Cheung, Ludwig Pettersson, Jonas Schneider, John Schulman,
  Jie Tang, and Wojciech Zaremba.
\newblock Openai gym.
\newblock \emph{arXiv preprint arXiv:1606.01540}, 2016.

\bibitem[{Brodeur} et~al.(2017){Brodeur}, {Perez}, {Anand}, {Golemo},
  {Celotti}, {Strub}, {Rouat}, {Larochelle}, and {Courville}]{HoME}
S.~{Brodeur}, E.~{Perez}, A.~{Anand}, F.~{Golemo}, L.~{Celotti}, F.~{Strub},
  J.~{Rouat}, H.~{Larochelle}, and A.~{Courville}.
\newblock {HoME: a Household Multimodal Environment}.
\newblock \emph{ArXiv e-prints}, November 2017.

\bibitem[Carlson et~al.(2010)Carlson, Betteridge, Kisiel, Settles, Hruschka~Jr,
  and Mitchell]{carlson2010toward}
Andrew Carlson, Justin Betteridge, Bryan Kisiel, Burr Settles, Estevam~R
  Hruschka~Jr, and Tom~M Mitchell.
\newblock Toward an architecture for never-ending language learning.
\newblock In \emph{AAAI}, volume~5, page~3. Atlanta, 2010.

\bibitem[{Chang} et~al.(2017){Chang}, {Dai}, {Funkhouser}, {Halber},
  {Nie{\ss}ner}, {Savva}, {Song}, {Zeng}, and {Zhang}]{matterport}
A.~{Chang}, A.~{Dai}, T.~{Funkhouser}, M.~{Halber}, M.~{Nie{\ss}ner},
  M.~{Savva}, S.~{Song}, A.~{Zeng}, and Y.~{Zhang}.
\newblock {Matterport3D: Learning from RGB-D Data in Indoor Environments}.
\newblock \emph{ArXiv e-prints}, September 2017.

\bibitem[Das et~al.(2018)Das, Datta, Gkioxari, Lee, Parikh, and
  Batra]{embodiedqa}
Abhishek Das, Samyak Datta, Georgia Gkioxari, Stefan Lee, Devi Parikh, and
  Dhruv Batra.
\newblock {E}mbodied {Q}uestion {A}nswering.
\newblock In \emph{Proceedings of the IEEE Conference on Computer Vision and
  Pattern Recognition (CVPR)}, 2018.

\bibitem[Duncan(2011)]{duncan2011minecraft}
Sean~C Duncan.
\newblock Minecraft, beyond construction and survival.
\newblock \emph{Well Played: a journal on video games, value and meaning},
  1\penalty0 (1):\penalty0 1--22, 2011.

\bibitem[Kempka et~al.(2016)Kempka, Wydmuch, Runc, Toczek, and
  Ja{\'s}kowski]{kempka2016vizdoom}
Micha{\l} Kempka, Marek Wydmuch, Grzegorz Runc, Jakub Toczek, and Wojciech
  Ja{\'s}kowski.
\newblock Vizdoom: A doom-based ai research platform for visual reinforcement
  learning.
\newblock In \emph{Computational Intelligence and Games (CIG), 2016 IEEE
  Conference on}, pages 1--8. IEEE, 2016.

\bibitem[Kiefer and Trumpp(2012)]{neuro_virtual_embodiment}
Markus Kiefer and Natalie~M. Trumpp.
\newblock Embodiment theory and education: The foundations of cognition in
  perception and action.
\newblock \emph{Trends in Neuroscience and Education}, 1\penalty0 (1):\penalty0
  15 -- 20, 2012.
\newblock ISSN 2211-9493.
\newblock \doi{https://doi.org/10.1016/j.tine.2012.07.002}.
\newblock URL
  \url{http://www.sciencedirect.com/science/article/pii/S221194931200004X}.

\bibitem[Koenig and Howard(2004)]{koenig2004design}
Nathan~P Koenig and Andrew Howard.
\newblock Design and use paradigms for gazebo, an open-source multi-robot
  simulator.
\newblock In \emph{IROS}, volume~4, pages 2149--2154. Citeseer, 2004.

\bibitem[Machado et~al.(2018)Machado, Bellemare, Talvitie, Veness, Hausknecht,
  and Bowling]{machado2018revisiting}
Marlos~C Machado, Marc~G Bellemare, Erik Talvitie, Joel Veness, Matthew
  Hausknecht, and Michael Bowling.
\newblock Revisiting the arcade learning environment: Evaluation protocols and
  open problems for general agents.
\newblock \emph{Journal of Artificial Intelligence Research}, 61:\penalty0
  523--562, 2018.

\bibitem[Mahmood et~al.(2018)Mahmood, Korenkevych, Vasan, Ma, and
  Bergstra]{mahmood2018benchmarking}
A~Rupam Mahmood, Dmytro Korenkevych, Gautham Vasan, William Ma, and James
  Bergstra.
\newblock Benchmarking reinforcement learning algorithms on real-world robots.
\newblock \emph{arXiv preprint arXiv:1809.07731}, 2018.

\bibitem[Mnih et~al.(2015)Mnih, Kavukcuoglu, Silver, Rusu, Veness, Bellemare,
  Graves, Riedmiller, Fidjeland, Ostrovski, et~al.]{mnih2015human}
Volodymyr Mnih, Koray Kavukcuoglu, David Silver, Andrei~A Rusu, Joel Veness,
  Marc~G Bellemare, Alex Graves, Martin Riedmiller, Andreas~K Fidjeland, Georg
  Ostrovski, et~al.
\newblock Human-level control through deep reinforcement learning.
\newblock \emph{Nature}, 518\penalty0 (7540):\penalty0 529, 2015.

\bibitem[{Puig} et~al.(2018){Puig}, {Ra}, {Boben}, {Li}, {Wang}, {Fidler}, and
  {Torralba}]{virtual_home}
X.~{Puig}, K.~{Ra}, M.~{Boben}, J.~{Li}, T.~{Wang}, S.~{Fidler}, and
  A.~{Torralba}.
\newblock {VirtualHome: Simulating Household Activities via Programs}.
\newblock \emph{ArXiv e-prints}, June 2018.

\bibitem[Puig et~al.(2018)Puig, Ra, Boben, Li, Wang, Fidler, and
  Torralba]{virtualhome}
Xavier Puig, Kevin Ra, Marko Boben, Jiaman Li, Tingwu Wang, Sanja Fidler, and
  Antonio Torralba.
\newblock Virtualhome: Simulating household activities via programs.
\newblock In \emph{CVPR}, 2018.

\bibitem[Ring(1997)]{ring1997child}
Mark~B Ring.
\newblock Child: A first step towards continual learning.
\newblock \emph{Machine Learning}, 28\penalty0 (1):\penalty0 77--104, 1997.

\bibitem[{Savva} et~al.(2017){Savva}, {Chang}, {Dosovitskiy}, {Funkhouser}, and
  {Koltun}]{Minos}
M.~{Savva}, A.~X. {Chang}, A.~{Dosovitskiy}, T.~{Funkhouser}, and V.~{Koltun}.
\newblock {MINOS: Multimodal Indoor Simulator for Navigation in Complex
  Environments}.
\newblock \emph{ArXiv e-prints}, December 2017.

\bibitem[Silver et~al.(2013)Silver, Yang, and Li]{silver2013lifelong}
Daniel~L Silver, Qiang Yang, and Lianghao Li.
\newblock Lifelong machine learning systems: Beyond learning algorithms.
\newblock In \emph{AAAI Spring Symposium: Lifelong Machine Learning},
  volume~13, page~05, 2013.

\bibitem[Silver et~al.(2016)Silver, Huang, Maddison, Guez, Sifre, Van
  Den~Driessche, Schrittwieser, Antonoglou, Panneershelvam, Lanctot,
  et~al.]{silver2016mastering}
David Silver, Aja Huang, Chris~J Maddison, Arthur Guez, Laurent Sifre, George
  Van Den~Driessche, Julian Schrittwieser, Ioannis Antonoglou, Veda
  Panneershelvam, Marc Lanctot, et~al.
\newblock Mastering the game of go with deep neural networks and tree search.
\newblock \emph{nature}, 529\penalty0 (7587):\penalty0 484, 2016.

\bibitem[Solomonoff(1989)]{solomonoff1989system}
Ray~J Solomonoff.
\newblock A system for incremental learning based on algorithmic probability.
\newblock In \emph{Proceedings of the Sixth Israeli Conference on Artificial
  Intelligence, Computer Vision and Pattern Recognition}, pages 515--527, 1989.

\bibitem[Song et~al.(2017)Song, Yu, Zeng, Chang, Savva, and
  Funkhouser]{suncgsong2016}
Shuran Song, Fisher Yu, Andy Zeng, Angel~X Chang, Manolis Savva, and Thomas
  Funkhouser.
\newblock Semantic scene completion from a single depth image.
\newblock \emph{IEEE Conference on Computer Vision and Pattern Recognition},
  2017.

\bibitem[Tessler et~al.(2017)Tessler, Givony, Zahavy, Mankowitz, and
  Mannor]{tessler2017deep}
Chen Tessler, Shahar Givony, Tom Zahavy, Daniel~J Mankowitz, and Shie Mannor.
\newblock A deep hierarchical approach to lifelong learning in minecraft.
\newblock In \emph{AAAI}, volume~3, page~6, 2017.

\bibitem[Thrun(1996)]{thrun1996explanation}
Sebastian Thrun.
\newblock \emph{Explanation-based neural network learning: A lifelong learning
  approach}, volume 357.
\newblock Springer Science \& Business Media, 1996.

\bibitem[{Wu} et~al.(2018){Wu}, {Wu}, {Gkioxari}, and {Tian}]{House3D}
Y.~{Wu}, Y.~{Wu}, G.~{Gkioxari}, and Y.~{Tian}.
\newblock {Building Generalizable Agents with a Realistic and Rich 3D
  Environment}.
\newblock \emph{ArXiv e-prints}, January 2018.

\bibitem[{Yan} et~al.(2018){Yan}, {Misra}, {Bennnett}, {Walsman}, {Bisk}, and
  {Artzi}]{CHALET}
C.~{Yan}, D.~{Misra}, A.~{Bennnett}, A.~{Walsman}, Y.~{Bisk}, and Y.~{Artzi}.
\newblock {CHALET: Cornell House Agent Learning Environment}.
\newblock \emph{ArXiv e-prints}, January 2018.

\bibitem[Zhu et~al.(2017)Zhu, Mottaghi, Kolve, Lim, Gupta, Fei-Fei, and
  Farhadi]{AI2-THOR}
Yuke Zhu, Roozbeh Mottaghi, Eric Kolve, Joseph~J Lim, Abhinav Gupta,
  Li~Fei-Fei, and Ali Farhadi.
\newblock Target-driven visual navigation in indoor scenes using deep
  reinforcement learning.
\newblock In \emph{Robotics and Automation (ICRA), 2017 IEEE International
  Conference on}, pages 3357--3364. IEEE, 2017.

\end{thebibliography}
\bibliographystyle{plainnat}

\end{document}